\documentclass[11pt,a4paper]{article}

\usepackage[hyperref]{acl2018}

\usepackage{times}
\usepackage{latexsym}
\usepackage{url}

\usepackage{graphicx}
\usepackage{amsmath}
\usepackage{array,multirow}
\usepackage{booktabs}
\usepackage{color}
\usepackage{textcomp}

\aclfinalcopy % Uncomment this line for the final submission

\title{Characterizing Variation in Crowd-Sourced Data for Training Neural Language Generators to Produce Stylistically Varied Outputs}

\author{Juraj Juraska \and Marilyn Walker \\
Natural Language and Dialogue Systems Lab \\
University of California, Santa Cruz \\
{\tt \{jjuraska,mawalker\}@ucsc.edu} \\}

\date{}

\begin{document}

\maketitle

\begin{abstract}

One of the biggest challenges of end-to-end language generation from meaning representations in dialogue systems is making the outputs more natural and varied. Here we take a large corpus of 50K crowd-sourced utterances in the restaurant domain and develop text analysis methods that systematically characterize types of sentences in the training data. We then automatically label the training data to allow us to conduct two kinds of experiments with a neural generator. First, we test the effect of training the system with different stylistic partitions and quantify the effect of smaller, but more stylistically controlled training data. Second, we propose a method of labeling the style variants during training, and show that we can modify the style of the generated utterances using our stylistic labels. We contrast and compare these methods that can be used with any existing large corpus, showing how they vary in terms of semantic quality and stylistic control.

\end{abstract}

\section{Introduction}
\label{sec:intro}

Dialogue systems have become one of the key applications in natural language processing, but there are still many ways in which these systems can be improved. One obvious possible improvement is in the system's language generation to make it more natural and more varied. Both a benefit and a challenge of neural natural language generation (NLG) models is that they are very good at reducing noise in the training data. When they are trained on a sufficiently large dataset, they learn to generalize and become capable of applying the acquired knowledge to unseen inputs. The more data the models are trained on, the more robust they become, which minimizes the effect of noise in the data on their learning. However, the higher amount of training data can also drown out interesting stylistic features and variations that may not be very frequent in the data. In other words, the model, being statistical, will prefer producing the most common sentence structures, i.e. those which it observed most frequently in the training data and is thus most confident about.

In our work, we consider language generators whose inputs are  structured \emph{meaning representations}~(MRs) describing a list of key concepts to be conveyed to the human user during the dialogue. Each piece of information is represented by a slot-value pair, where the \emph{slot} identifies the type of information and the \emph{value} is the corresponding content. A language generator must produce a syntactically and semantically correct utterance from a given MR. The \emph{utterance} should express all the information contained in the MR, in a natural and conversational way. Table~\ref{table:utt_MR_example} shows an example MR for a restaurant called \emph{The Waterman} paired with two (of many) possible output utterances, the first of which might be considered stylistically interesting, since the name of the restaurant follows some aspects of the description and contains a concession, while the second example might be considered as more stylistically conventional.

\begin{table}
	\begin{center}
    \begin{tabular}{>{\centering\arraybackslash} m{0.15\linewidth} m{0.75\linewidth}}
    	\toprule
    	\textbf{MR} & name [\textbf{The Waterman}], food [\textbf{English}], priceRange [\textbf{cheap}], customer rating [\textbf{low}], area [\textbf{city centre}], familyFriendly [\textbf{yes}]\\
        \midrule
        \textbf{Utt. \#1} & There is a \textbf{cheap}, \textbf{family-friendly} restaurant in the \textbf{city centre}, called \textbf{The Waterman}. It serves \textbf{English} food, but received a \textbf{low} rating by customers.\\
        \midrule
        \textbf{Utt. \#2} & \textbf{The Waterman} is a \textbf{family-friendly} restaurant in the \textbf{city centre}. It serves \textbf{English} food at a \textbf{cheap} price. It has a \textbf{low} customer rating.\\
        \bottomrule
    \end{tabular}
  	\end{center}
    \caption{Example of a meaning representation and two corresponding utterances of different styles.}
   	\label{table:utt_MR_example}
\end{table}

Recently, the size of training corpora for NLG has become larger, and these same corpora have begun to manifest interesting stylistic variations. Here we start from the recently released E2E dataset~\cite{novikova2017e2e} with nearly 50K samples of crowd-sourced utterances in the restaurant domain provided as part of the E2E NLG Challenge\footnote{http://www.macs.hw.ac.uk/InteractionLab/E2E/}. We first develop text analysis methods that systematically characterize types of sentences in the training data. We then automatically label the training data to allow us to conduct two kinds of experiments with a neural language generator: (1) we test the effect of training the system with different stylistic partitions and quantify the effect of smaller, but more stylistically controlled training data; (2) we propose a method of labeling the style variants during training, and show that we can modify the style of the output using our stylistic labels. We contrast these methods, showing how they vary in terms of semantic quality and stylistic control. These methods promise to be usable with any sufficiently large corpus as a simple way of producing stylistic variation.

\section{Related Work}
\label{subsec:related_work}

The restaurant domain has always been the domain of choice for NLG tasks in dialogue systems \cite{stent2004trainable,gavsic2008training,mairesse2010phrase,howcroft2013enhancing}, as it offers a good combination of structured information availability, expression complexity, and ease of incorporation into conversation. Hence, even the more recent neural models for NLG continue to be tested primarily on data in this domain \cite{wen2015semantically,duvsek2016sequence,nayak2017plan}. These tend to focus solely on syntactic and semantic correctness of the generated utterances, nevertheless, there have also been recent efforts to collect training data for NLG with emphasis on stylistic variation \cite{nayak2017plan,novikova2017e2e_dataset,oraby2017harvesting}.

While there is previous work on stylistic variation in NLG \cite{paiva2004framework,mairesse2007personage}, this work did not use crowd-sourced utterances for training. More recent work in neural NLG that explores stylistic control has not needed to control semantic correctness, or examined the interaction between semantic correctness and stylistic variation \cite{sennrich2016controlling,ficler2017controlling}. Also related is the work of \citet{niu2017discovering} that analyzes how dense word embeddings capture style variations, \citet{kabbara2016stylistic} who explore the ability of neural NLG systems to transfer style without the need for parallel corpora, which are difficult to collect \cite{rao2018dear}, while \citet{li2018delete} use a simple delete-and-retrieve method also without alignment to outperform adversarial methods in style transfer. Finally, \citet{oraby2018controlling} propose two different methods that give neural generators control over the language style, corresponding to the Big Five personalities, while maintaining semantic fidelity of the generated utterances.

To our knowledge, there is no previous work exploring the use of and utility of stylistic selection for controlling stylistic variation in NLG from structured MRs. This may be either because there have not been sufficiently large corpora in a particular domain, or because it is surprising, as we show, that relatively small corpora (2000 samples) whose style is controlled can be used to train a neural generator to achieve high semantic correctness while producing stylistic variation.

\section{Dataset}
\label{subsec:dataset}

We perform the stylistic selection on the E2E dataset~\cite{novikova2017e2e}. It is by far the largest dataset available for task-oriented language generation in the restaurant domain. It offers almost 10 times more data than the San Francisco restaurant dataset~\cite{wen2015semantically}, which had frequently been used for NLG benchmarks. This significant increase in size allows successful training of neural models on smaller subsets of the dataset. Careful selection of the training subset can be used to influence the style of the utterances produced by the model, as we show in this paper.

The human reference utterances were collected using pictures as the source of information, which was shown to inspire more natural utterances compared to textual MRs~\cite{novikova2016crowd}. The reference utterances in the E2E dataset exhibit superior lexical richness and syntactic variation, including more complex discourse phenomena. It aims to provide higher-quality training data for end-to-end NLG systems to learn to produce better phrased and more naturally sounding utterances.

\begin{table}
 	\centering
  	\begin{tabular}{l r r}
    	\toprule
    	& \textbf{Samples}	& \textbf{Unique MRs}	\\
    	\midrule
    	\textbf{Training}	& 42,061	& 4,862 \\
    	\textbf{Validation}	& 4,672		& 547 \\
    	\textbf{Test}		& 630		& 630 \\
    	\midrule
    	\textbf{Total}		& 47,363	& 6,039 \\
    	\bottomrule
  	\end{tabular}\\
  	\caption{Number of samples vs. unique meaning representations in the training, validation and test set of the E2E dataset.}
  	\label{table:dataset_stats}
\end{table}

Although the E2E dataset contains a large number of samples, each MR is associated on average with more than 8 different reference utterances, effectively supplying almost 5K unique MRs in the training set (Table~\ref{table:dataset_stats}). It thus offers multiple alternative ways of expressing the same information in an utterance, which the model can learn. We take advantage of this aspect of the dataset when selecting the subset of samples for training with a particular purpose of stylistic variation.

The dataset contains 8 different slot types, which are fairly equally distributed in the dataset. Each MR comprises 3 to 8 slots, whereas the majority of MRs consist of 5 and 6 slots. Even though most of the MRs contain many slots, the majority of the corresponding human utterances, however, consist of one or two sentences only (Table~\ref{table:sentence_average}), suggesting a reasonably high level of sentence complexity in the references.

\begin{table}
	\small
	\centering
 	\begin{tabular}{l r r r r r r}
		\toprule
   		\textbf{Slots}		& 3	& 4	& 5	& 6	& 7	& 8 \\
   		\midrule
   		\textbf{Sentences}	& 1.09 & 1.23 & 1.41 & 1.65 & 1.84 & 1.92 \\
   		\textbf{Proportion}	& 5\% & 18\% & 32\% & 28\% & 14\% & 3\% \\
   		\bottomrule
	\end{tabular}\\
 	\caption{Average number of sentences in the reference utterance for a given number of slots in the corresponding MR, along with the proportion of MRs with specific slot counts.}
 	\label{table:sentence_average}
\end{table}

\section{Stylistic Selection}
\label{sec:stylistic_selection}

We note that the E2E dataset is significantly larger than what is  needed for a neural model to learn to produce correct utterances in this domain. Thus, we seek a way to help the model learn more than just to be correct. We strive to achieve higher stylistic diversity of the utterances generated by the model through stylistic selection of the training samples. We start by characterizing variation in the crowd-sourced dataset and detect what opportunities it offers for the model to learn more advanced sentence structures. Table~\ref{table:discourse_phenomena} illustrates some of the stylistic variation that we observe, which we describe in more detail below. We then judge the level of desirability of specific discourse phenomena in our context, and devise rules based on the parse tree to extract the samples that manifest those stylistic phenomena. This gives us the ability to create subsets of the samples with an arbitrary combination of stylistic features that we are interested in. We then explore the extent to which we can make the model's output demonstrate these stylistic features.

\subsection{Stylistic Variation in the Dataset}
\label{subsec:stylistic_variation}

\begin{table*}
	\begin{center}
    \begin{tabular}{ m{0.18\linewidth} m{0.77\linewidth} }
    	\toprule
        \textbf{Category} & \textbf{Utterance} \\
    	\midrule
        Aggregation & Located in the city centre is a family-friendly coffee shop called Fitzbillies. It is \textbf{both inexpensive and highly rated}. \\
    	\midrule
        Contrast & The Rice Boat is a Chinese restaurant in the riverside area. It has a customer rating of 5 out of 5 \textbf{but} is not family friendly. \\
    	\midrule
        Fronting & \textbf{With a 1 out of 5 rating} Midsummer House serves Italian cuisine in the high price range, found not far from All Bar One. \\
    	\midrule
        Subordination & Wildwood pub is serving 5 star food \textbf{while keeping their prices low}. \\
    	\midrule
        Exist. clause & In the city center, \textbf{there} is an average priced, non-family-friendly, Japanese restaurant called Alimentum. \\
    	\midrule
        Imperative/modal & In Riverside, \textbf{you'll} find Fitzbillies. It is a passable, affordable coffee shop which interestingly serves Chinese food. \textbf{Don't bring} your family though. \\
    	\bottomrule
    \end{tabular}
  	\end{center}
    \caption{Examples of the categories of discourse phenomena extracted from the utterances in the E2E dataset.}
   	\label{table:discourse_phenomena}
\end{table*}

This section gives an overview of different discourse phenomena in the E2E dataset that we consider relevant in the context of a task-oriented dialogue in the restaurant domain. The majority of these would, however, generalize to other domains too, and so the extraction rules we have implemented can be widely used in task-oriented language generators. We split the sentence features in the following six categories. An example of each is given in in Table~\ref{table:discourse_phenomena}:
\begin{itemize}
    \item \textbf{Aggregation:} Discourse phenomena grouping information together in a more concise way. This includes specifiers such as ``both'' or ``also'', as well as apposition and gerunds. Another type of aggregation uses the same quantitative adjective for characterizing multiple different qualities (such as ``It has a \emph{low customer rating and price range}.'').
    
    Note that some of the following categories contain other markers that also represent aggregation.
	\item \textbf{Contrast:} Connectors and adverbs expressing concession or contrast between two or more qualities, such as ``but'', ``despite'', ``however'', or ``yet''.
    \item \textbf{Fronting:} Fronted adjective, verb and prepositional phrases, typically highlighting qualities of the eatery before its name is given.
    
    In this category we also include specificational copular constructions, which are formulations with inverted predication around a copula, bringing a particular quality of the eatery in the front (e.g. ``\emph{A family friendly option is} The Rice Boat.'').
    \item \textbf{Subordination:} Clauses introduced by a subordinating conjunction (such as ``if'' or ``while''), or by a relative pronoun (such as ``whose'' or ``that'').
    \item \textbf{Existential clause:} Sentences formulated using the expletive ``there''.
    \item \textbf{Imperative and modal verb:} Sentences involving a verb in the imperative form or a modal verb, making the utterance sound more personal and interactive.
\end{itemize}

% TODO: noise in the dataset in the form of grammatically incorrect utterance, plus our heuristic methods are not exact

\subsection{Discourse Marker Weighting}

Many human-produced utterances, naturally, contain multiple of the discourse phenomena described in Section~\ref{subsec:stylistic_variation}. Such utterances are preferred to those only containing a single discourse phenomenon of interest, especially if it is a common one, such as the existential clause. We therefore devise a weighting schema for different groups of discourse markers, whose purpose is to represent the markers' general desirability in the output utterances, as well as to counteract the sparsity of some of the markers compared to others. In other words, the weighting is supposed to ensure all the most desirable utterances are picked from the training set during the selection, but some that only contain less interesting (and typically more prevalent) discourse phenomena would be omitted in favor of the more complex ones. Our reasoning behind that is that the greater the proportion of the most desirable discourse phenomena in the stylistically selected training set, the more confidently the model is expected to generate utterances in which they are present.

For an illustration, let us assume there are eight different reference utterances for an MR. All of them will be scored based on the discourse markers they contain, but only those that score above a certain threshold will be selected, while the rest will be ignored. The purpose of that is to encourage the model to learn to use, say, a contrastive phrase if there is an opportunity for it in the MR, and not be distracted by other possible realizations of the same MR, which are not as elegant (such as the example utterance \#1 vs. \#2 in Table~\ref{table:utt_MR_example}). Thus, we can set the weighting schema in such a way that sentences containing only, for example, ``which'' or an existential clause,  will not be picked. However, if there is no high scoring utterance for an MR, the utterance with the highest score is picked so that the model would not miss an opportunity to learn from any MR samples.

\begin{table*}
	\centering
    \begin{tabular}{l l r r}
      \toprule
      \textbf{Category}	& \textbf{Subset of markers}	& \textbf{Proportion}	& \textbf{Weight}	\\
      \midrule
      \multirow{3}{*}{Aggregation}	& ``also, both, neither,...'', quantitative adjectives	& 1.8\%	& 3 \\
                                      & apposition	& 4.6\%	& 2 \\
                                      & gerund	& 11.2\%	& 2 \\
      \midrule
      Contrast		& ``but, however, despite, although,...''	& 5.4\%	& 3 \\
      \midrule
      Fronting		& fronted adjective/prepositional/verb clause	& 14.5\%	& 2 \\
      \midrule
      \multirow{2}{*}{Subordination}	& subordinating conj.	& 2.9\%	& 2	\\
                                      & relative pronouns	& 19.3\%	& 1	\\
      \midrule
      Existential clause		& expletive ``there''	& 10.0\%	& 1 \\
      \midrule
      \multirow{2}{*}{Imperative/modal}		& imperative	& 1.0\%	& 2 \\
                                              & modal verb	& 4.1\%	& 2 \\
      \bottomrule
  	\end{tabular}\\
  	\caption{The weighting schema for different discourse markers for each introduced category of discourse phenomena. For each set of markers we indicate the heuristically determined proportion of reference utterances in the training set they appear in.}
    \label{table:weighting_schema}
\end{table*}

Our final weighting schema is specified in Table~\ref{table:weighting_schema}. When there are discourse markers from multiple subsets present in the utterance, the weights are accumulated. It is then the total weight that is used to determine whether the utterance satisfies the stylistic threshold or should be eliminated.

The weights have been determined through a combination of the discourse markers' frequency in the dataset, their intra-category variation, as well as their general desirability in the particular domain of our task. The weights can be easily adjusted for any new domain according to the above, or any other factors. As an example, another such factor could be the length of the utterance. We have experimented with a length penalty, i.e. giving an utterance that contains a verb in gerund form as the only advanced construct, but that is composed of three sentences, a lower score than a short one-sentence utterance with a gerund verb. However, we did not find the use of this extra coefficient helpful in our domain, as it resulted in eliminating a significant proportion of desirable utterances too.

% I experimented with dividing the score by the number of sentences, as well as subtracting a constant multiplied by the number of sentences from the score, along with different score thresholds.

\section{Data Annotation}
\label{sec:data_annotation}

% TODO: motivate the data annotation

\subsection{Contrastive Relation}

One of the discourse phenomena whose actualization could benefit from explicit indication of when it should be applied, is the contrastive relation between two (or more) slot realizations in the utterance. There are several reasons why such a comparison of specific slots would be desired in the restaurant domain. One of them is to provide emphasis that one attribute is positive, whereas the other is negative. Another natural reason in dialogue systems could be to indicate that the closest match to the user's query that was found is a restaurant that does not satisfy one of the requested criteria. A third instance is when the value of one attribute creates the expectation of a particular value of another attribute, but the latter has in reality the opposite value.

Some of the above could presumably be learned by the model if sufficient training data was available. However, they involve fairly complex sentence constructs with various potentially confusing rules for the neural network. The slightly more than 2K samples with a contrasting relation can be drowned among the thousands of other samples in the E2E dataset, meaning that it is difficult for the learned model to produce them.  

Hence, we augment the input given to the model with the information about which slots should be put into a contrastive relation. We hypothesize that this explicit indication will help the model to learn to apply contrasting significantly more easily despite the small proportion of training samples displaying the property.

In order to extract the information as exactly as possible from the training utterance, we use a heuristic slot aligner \cite{juraska2018slug2slug} to identify two slots that are in a contrastive relation. For the relation we only consider the two scalar slots (\emph{price range} and \emph{customer rating}), plus the boolean slot \emph{family friendly}. Whenever a contrastive relation appears to the aligner to involve a slot other than the above three, we discard it as an undesirable utterance formulation. Depending on the values of the two identified slots, we assign the sample either of the following labels:
\begin{itemize}
	\item \textbf{Contrast:} If the slots have different values on the 3-level positivity scale they can be mapped to (the \emph{family friendly} slot is only mapped to levels $\{1, 3\}$). An example would be \emph{customer rating} being ``low'' ($\rightarrow 1$) and \emph{family friendly} having value ``yes'' ($\rightarrow 3$).
	\item \textbf{Concession:} If the slots have an equivalent value. For instance, \emph{customer rating} being ``5 out of 5'' ($\rightarrow 3$) and \emph{price range} having value ``cheap'' ($\rightarrow 3$).
\end{itemize}

The label is added in the form of a new auxiliary slot in the MR, containing the names of the two corresponding slots as its value, such as $<$\texttt{contrast}$>$ \texttt{[priceRange customer\_rating]}.

We observed instances in the dataset that, semantically, can be classified neither as contrast nor as concession, but using our above rules, they would be considered a concession. An example of such a reference utterance is: ``Strada is a low price restaurant located near Rainbow Vegetarian Caf\'e serving English food with a \emph{low customer rating but not family-friendly}.'' Notice that the emphasized part of the utterance contains a questionable use of the word ``but'', as both of the attributes of the restaurant (customer rating and family-friendliness) are negative. Such utterances were, however, scarce, and thus we considered them as an acceptable noise.

\subsection{Emphasis}
\label{subsec:emphasis}

Another utterance property that might in practice be desired to be indicated explicitly and, in that way, enforced in the output utterance, is emphasis. Through fronting discourse phenomena, such as specificational copular constructions or fronted prepositional phrases, certain information about the subject can be emphasized at the beginning of the utterance.

\begin{table}
	\begin{center}
    \begin{tabular}{ >{\centering\arraybackslash} m{0.2\linewidth} m{0.7\linewidth} }
    	\toprule
        \textbf{User query} & Is there a family-friendly Indian restaurant nearby? \\
    	\midrule
        \textbf{Response with no emphasis} & \emph{The Rice Boat} in city centre near Express by Holiday Inn is serving Indian food at a high price. It is \textbf{family-friendly} and received a customer rating of 1 out of 5. \\
    	\midrule
        \textbf{Response with emphasis} & A \textbf{family-friendly} option is \emph{The Rice Boat}. This Indian cuisine is priced on the higher end and has a rating of 1 out of 5. They are located near Express by Holiday Inn in the city centre. \\
    	\bottomrule
    \end{tabular}
  	\end{center}
    \caption{An example of emphasizing the information about family-friendliness in an utterance conveying the same content.}
   	\label{table:emphasis_example}
\end{table}

\begin{table*}[ht]
	\begin{center}
    \begin{tabular}{ m{0.15\linewidth} m{0.8\linewidth} }
    	\toprule
        \textbf{MR} & name [Wildwood], \emph{eatType [coffee shop]}, \emph{food [English]}, priceRange [moderate], customer rating [1 out of 5], \emph{near [Ranch]} \\
    	\midrule
        \textbf{Reference} & A low rated English style coffee shop around Ranch called Wildwood has moderately priced food. \\
    	\midrule
        \textbf{No emph.} & Wildwood is a coffee shop providing English food in the moderate price range. It is located near Ranch. \\
    	\midrule
        \emph{\textbf{With emph.}} & There is an English coffee shop near Ranch called Wildwood. It has a moderate price range and a customer rating of 1 out of 5. \\
    	\bottomrule
    \end{tabular}
  	\end{center}
    \caption{Examples of generated utterances with or without an explicit emphasis annotation.}
   	\label{table:emph_annot_ouputs}
\end{table*}

This could be used to make the dialogue system's responses sound more context-aware and thus natural. Consider the following example in the restaurant domain. Assume the user asks the agent for a recommendation of a family-friendly Indian restaurant (see Table~\ref{table:emphasis_example}). Considering they have explicitly specified the ``family-friendly'' requirement in the query, it is arguably more natural for the response utterance to be in the form of the second response example in the table rather than the first.

We argue that the order of the information given in the response matters and should not be entirely random. That motivated us to identify instances in the training set where some information about the restaurant is provided in the utterance before its name. In order to do so, and to extract the information about which slot(s) the segment of the utterance represents, we employ the heuristic slot aligner once again. Subsequently, we augment the corresponding input to the model with additional $<$\texttt{emph}$>$ tokens before the slots that should be emphasized in the output utterance. This additional indication will give the model an incentive to learn to realize such slots at the beginning of the utterance when desired. From the perspective of the dialogue manager in a dialogue system, it simply needs to indicate slots to emphasize along with the generated MR whenever applicable.

\section{Evaluation}
\label{sec:evaluation}

\subsection{Experimental Setup}

For our sequence-to-sequence NLG model we use the standard encoder-decoder \cite{cho} architecture equipped with an attention mechanism as defined in~\citet{bahdanau2015neural}. The samples are delexicalized before being fed into the model as input, so as to enhance the ability of the model to generalize the learned concepts to unseen MRs. We only delexicalize categorical slots whose values always propagate verbatim from the MR to the utterance. The corresponding values in the input MR get thus replaced with placeholder tokens for which the values from the original MR are eventually substituted in the output utterance as a part of post-processing.

We use a 4-layer bidirectional LSTM~\cite{hochreiter1997long} encoder and a 4-layer LSTM decoder, both with 512 cells per layer. During inference time, we use beam search with the beam width of 10 and length normalization of the beams as defined in~\citet{wu2016google}. The length penalty that we determined was providing the best results on the E2E dataset was 0.6. The beam search candidates are reranked using a heuristic slot aligner as described in~\citet{juraska2018slug2slug}, and the top candidate is returned as the final utterance.

\subsection{Style Subsets}

In the initial experiments, we trained the model on the reduced training set, which only contains the utterances filtered out based on the weighting schema defined in Table~\ref{table:weighting_schema}. Setting the threshold to 2, we obtained a training set of 17.5K samples, which is approximately 40\% of the original training set. Although this reduced training set had a higher concentration of more desirable reference utterances, the dataset turned out to be still too general with most of the rare discourse phenomena drowned out. However, many of them, including contrast, apposition and fronting, appeared multiple times in the generated utterances in the test set, which was not the case for a model trained on the full training set.

Therefore, our next step was to verify whether our model is capable of learning all the concepts of the discourse phenomena individually and apply them in generated utterances. To that end we repeatedly trained the model on subsets of the E2E dataset, each containing only samples with a specific group of discourse markers as listed in the second column of Table~\ref{table:weighting_schema}. We then evaluated the outputs on the correspondingly reduced test set, using the same method we used for identifying samples with specific discourse markers, as described in Section~\ref{subsec:stylistic_variation}. In other words, we identified what proportion of the generated utterances did exhibit the desired discourse phenomenon.

The results show that the model is indeed able to learn how to produce various advanced sentence structures that are, moreover, syntactically correct despite being trained on a rather small training set (in certain cases $<$2K samples). In all of the experiments, 97--100\% of the generated utterances conformed to the style the model was trained to produce. Any occasional incoherence that we observed (e.g. ``It has a high customer rating, but \emph{are} not kid friendly.'') was actually picked up from poor reference utterances in the training set. The only exception in the syntactic correctness was the \emph{Imperative/modal} category. Since this is one of the least represented categories among the six, and due to the particularly high complexity and diversity of the utterances, the model trained exclusively on the samples in this category generated a significant proportion of slightly incoherent utterances.

% TODO: mention that the most frequently present feature dominates the outputs (e.g. "that" among subordinating clauses)

\subsection{Data Annotation}

The first set of experiments we performed with the data annotation involved explicit indication of emphasis in the input (see Section~\ref{subsec:emphasis}). As the results in Table~\ref{table:data_annot_err} show, the model trained on data with emphasis annotation reached an almost 98\% success rate of generating an utterance with the desired slots emphasized.\footnote{There were 3,309 slots across all the test MRs that were labeled as to-be-emphasized.} In order to get a better idea of the impact of the annotation, notice that the same model trained on non-annotated data does not produce a single utterance with emphasis. The latter model defaults to producing utterances in a rigid style, which always starts with the name of the restaurant (see Table~\ref{table:emph_annot_ouputs}).

We notice that the error rate of the slot realization rises (from 3.45\% to 5.82\%) when the annotation is introduced. Nevertheless, it is still lower than the error rate among the reference utterances in the test set, in which over 8\% of slots have missing mentions. Thus we find it acceptable considering the desired stylistic improvement of the output utterances.

\begin{table}
	\centering
  	\begin{tabular}{l r r}
      \toprule
      	& \textbf{Emph. realiz.}	& \textbf{Slot error rate} \\
      \midrule
      \textbf{Reference}	& 100.00\%	& 8.48\% \\
      \textbf{No emph.}		& 0.00\%	& 3.45\% \\
      \textbf{With emph.}	& 97.85\%	& 5.82\% \\
      \bottomrule
  	\end{tabular}
  	\caption{Comparison of the emphasis realization success rate and the slot realization error rate in the generated outputs using data annotation against the reference utterances, as well as the outputs of the same model trained on non-annotated data.}
  	\label{table:data_annot_err}
\end{table}

The experiments with contrastive relation annotation also show a significant impact of the added labels on the style of the output utterances produced by our model. However, the success rate of the realization of a contrast/concession formulation was only 49.12\%, and the slot realization error rate jumped up to 8.34\%. The contrast and concession discourse phenomena being syntactically more complex, and at the same time being less prevalent among the training utterances, it is understandable that it was more difficult for the model to learn how to use them properly.

\subsection{Aggregation}

One of the aggregation discourse markers that we identified in Section~\ref{subsec:stylistic_variation} as contributing to the stylistic variation in an interesting way is, unfortunately, very sparsely represented in the E2E dataset. It is the last aggregation type described in the category overview in Section~\ref{subsec:stylistic_variation}. Its scarcity in the training set would not make it feasible to train a successful neural model on the subset of the corresponding samples only.

Nevertheless, we analyze the potential for this aggregation in the training set. Since there are only two scalar slots in this dataset -- \emph{price range} and \emph{customer rating} -- we obtain the frequencies of their value combinations. Both of these take on values on a scale of 3, however, the values are different for each of the slots. Moreover, there are two sets of values for both slots throughout the dataset. We have observed, however, that the values between the two sets are used somewhat interchangeably in the utterances, e.g. ``low'' seems to be a valid expression of the ``less than \pounds20'' value of the \emph{price range} slot, and vice versa.

\begin{table}
	\centering
  	\begin{tabular}{c c r}
      \toprule
      \textbf{Price range}	& \textbf{Customer rating}	& \textbf{Frequency} \\
      \midrule
      less than \pounds20	& low 			& 2,153 \\
      \pounds20-25		& 3 out of 5 	& 919 \\
      moderate			& 3 out of 5 	& 1,282 \\
      more than \pounds30	& high 			& 1,329 \\
      more than \pounds30	& 5 out of 5 	& 921 \\
      \bottomrule
  	\end{tabular}
  	\caption{Combinations of the slot values for which aggregation would be feasible. Note that only the combinations with a non-zero frequency are listed.}
  	\label{table:aggregation_potential}
\end{table}

As can be seen in Table~\ref{table:aggregation_potential}, the potential for the aggregation is rather limited. Although the 6,604 samples in which a feasible value combination can be found corresponds to over 15\% of the training set, due to the values not matching exactly between the two slots, aggregation was not elicited in the utterances. Moreover, a high value in the \emph{customer rating} means it is a positive attribute, while a high value in the \emph{price range} slot indicates a negative attribute. We conjecture this might have also deterred the crowd-source workers who produced the utterances from aggregating the values together.

\section{Conclusion}

In this paper we have presented two different methods of giving a neural language generation system greater stylistic control. Our results indicate that the data annotation method has a significant impact on the model being able to learn how to use a specific style and sentence structures, without an unreasonable impact on the error rate. As our future work, we plan to utilize transfer learning in the style-subset method to improve the model's ability to apply various different styles at the same time, wherein we would also make further use of the weighting schema. Finally, these methods are a convenient way for achieving the goal of stylistic control when training a neural model with an arbitrary existing large corpus.

% TODO: Mention transfer learning as future work. The model could be trained on the more common samples, and then fine-tuned on the subset with more complex samples only.

%\section{Acknowledgements}

\bibliographystyle{acl_natbib}
\bibliography{references}

\end{document}